\documentclass[serif, twocolumn]{wiley-article}
\usepackage[super]{natbib}
\makeatletter
\renewcommand{\@biblabel}[1]{#1.}
\makeatother

\usepackage[utf8]{inputenc}
\usepackage{siunitx}
\usepackage{multirow} 
\usepackage{graphicx}
\usepackage{makecell}
\usepackage{float}
\usepackage{subcaption}
\usepackage[labelformat=parens,labelsep=quad,skip=3pt]{caption}
\usepackage{hyperref}
\usepackage{color, colortbl}
\usepackage{pdflscape}
\usepackage{rotating}
\usepackage{array}
\usepackage{mathtools, cuted}
\usepackage{bm}
\usepackage{relsize}

\definecolor{Gray}{gray}{0.8}

\graphicspath{{./images/}} 

\newcolumntype{P}[1]{>{\centering\arraybackslash}p{#1}}

\DeclareUnicodeCharacter{2212}{\textminus}
\usepackage[utf8]{inputenc}
\DeclareUnicodeCharacter{2010}{-}
\papertype{Original Article}
\paperfield{Journal Section}

\title{\relsize{-1}MultiCalib4DEB: A toolbox exploiting multimodal optimisation in Dynamic Energy Budget parameters calibration}

\author[1]{\relsize{-1}Juan Francisco Robles}
\author[1]{\relsize{-1}Manuel Chica}
\author[2,4]{\relsize{-1}Ram\'on Filgueira}
\author[2]{\relsize{-1}Antonio Ag\"{u}era}
\author[3]{\relsize{-1}Sergio Damas}

\affil[1]{{\relsize{-1}Department of Computer Science and Artificial Intelligence, Andalusian Research
Institute in Data Science and Computational Intelligence, DaSCI, University of
Granada, Granada, 18071, Spain}}
\affil[2]{{\relsize{-1}Department of Benthic resources, Norwegian Institute of Marine Research, Bergen, NO-5817, Norway}}
\affil[3]{{\relsize{-1}Department of Software Engineering, Andalusian Research
Institute in Data Science and Computational Intelligence, DaSCI, University of
Granada, Granada, 18071, Spain}}
\affil[4]{{\relsize{-1}Marine Affairs Program, Life Sciences Centre, Dalhousie University, 1459 Oxford Street, Halifax, Nova Scotia, B3H 4R2, Canada}}
\corraddress{{\relsize{-1}Juan Francisco Robles, Department of Computer Science and Artificial Intelligence, Andalusian Research Institute in Data Science and Computational Intelligence, DaSCI, University of
Granada, Granada, 18071, Spain}}
\corremail{jfrobles@ugr.es}
\fundinginfo{{\relsize{-1}This work was supported by the Spanish Ministry of Science, Innovation and Universities, the Andalusian Government, the University of Granada, the Spanish National Agency of Research (AEI), and European Regional Development Funds (ERDF) under grants AIMAR (A-TIC-284-UGR18), and SIMARK (PY18-4475).}}
\runningauthor{Robles et al.}
\begin{document}

\begin{frontmatter}
\maketitle

\begin{abstract}
\begin{enumerate}
    \item Calibration is a crucial step for the validation of computational models and a challenging task to accomplish.
    \item Dynamic Energy Budget (DEB) theory has experienced an exponential rise in the number of published papers, which in large part has been made possible by the DEBtool toolbox. Multimodal evolutionary optimisation could provide DEBtool with new capabilities, particularly relevant on the provisioning of equally optimal and diverse solutions.
    \item In this paper we present MultiCalib4DEB, a MATLAB toolbox directly integrated into the existing DEBtool toolbox, which uses multimodal evolutionary optimisation algorithms to find multiple global and local optimal and diverse calibration solutions for DEB models.
    \item MultiCalib4DEB adds powerful calibration mechanisms, statistical analysis, and visualisation methods to the DEBtool toolbox and provides a wide range of outputs, different calibration alternatives, and specific tools to strengthen the DEBtool calibration module and to aid DEBtool users to evaluate the performance of the calibration results.
\end{enumerate}

\end{abstract}
\end{frontmatter}

\section{Introduction}
\label{sec:introduction}
Dynamic Energy Budget (DEB) theory~\citep{kooijman2008, kooijman2010} has become a popular approach to appropriately describe individual’s bioenergetics throughout their life cycle. DEB theory has been extensively used to represent the metabolism of different species~\citep{sara2013, smallegange2017}. However, the need to calibrate a large set of parameters using many data sets simultaneously is an obstacle for the use of DEB models and one of the most challenging aspects of the modelling. Moreover, parameters to be calibrated for thousands of species and DEB parameter calibration has become a core task~\citep{lika2014}.

The creation of the Add-my-Pet project\footnote{\url{http://www.bio.vu.nl/thb/deb/deblab/add_my_pet/}} and the introduction of the DEBtool toolbox~\citep{DEBtool, martin2012} provided a convenient an accessible calibration framework and a vast library of functions and examples to use for parameter calibration and DEB models applications. DEBtool was one of the reasons behind the success of DEB theory as it is being constantly updated to improve the modelling experience~\citep{lika2011, lika2014, marques2018, marques2019}. For this reason, several authors have used the toolbox as the basis for the parameters calibration of the species they work with~\citep{sara2013}, proposed several methods to enhance both DEBtool calibration results~\citep{morais2019} and its validation~\citep{accolla2020}, and contributed to the Add-my-Pet project. Up to July 2021, around 3,000 species are available on the project database\footnote{\url{http://www.bio.vu.nl/thb/deb/deblab/add_my_pet/species_list.html}}. 

The existing calibration module of DEBtool uses a Nelder Mead (NM) simplex method~\citep{Nelder1965} for the calibration of the parameters. Although some authors have proposed procedures to carry out the calibration process in a more efficient way~\citep{morais2019}, the main problem with such methods is that they are not able to deal with the challenges of the multimodal search space of the DEB parameters calibration, such as the existence of different sets of calibrated parameter values that frequently yield similar fitting qualities~\citep{Chica17Ins}. Thus, the results that traditional calibration methods return are limited to a single solution. This fact limits the exploration of the multimodal search space and the alternatives and knowledge offered to the biological modeler. There are more sophisticated optimisation approaches in the field of operations research and artificial intelligence that better deal with these multimodal optimisation problems. 

The multimodal nature of the calibration problem and the existence of non-linear correlations between the set of parameters to be calibrated usually make approximate optimisation algorithms, such as evolutionary algorithms~\citep{back1996evolutionary}, the best approach to tackle different multimodal optimisation problems~\citep{Chica17Ins, Filgueira2020, Robles21}. Specifically, multimodal evolutionary algorithms (MMEAs) has been recognised as a powerful method to obtain diverse and high quality solutions in large and complex problems while improving the validation of the results in a reasonable time. MMEAs are able to return different optimal parameter configurations with similar fittings that are equally preferable to each other, providing additional insights for sensitivity analysis and about the model's robustness~\citep{Chica17Ins}. Thanks to the recent discovery of the boundaries of the parameter space of the DEB models~\citep{lika2014} and the application of filters, it is possible to improve the calibration of DEB models using MMEAs. These boundaries represent the values within which the calibration parameters can fluctuate, avoiding unfeasible results. 

In this work we present a calibration toolbox based on MMEAs~\citep{das2011}, called MultiCalib4DEB, which is directly integrated with the existing DEBtool and its calibration framework\footnote{\url{https://github.com/add-my-pet/DEBtool_M/tree/master/lib/MultiCalib4DEB}}. MultiCalib4DEB v.1.0 is available in GitHub\footnote{\url{https://github.com/JuanfranRobles/MultiCalib4DEB}}  under the GNU license. The presented toolbox adds the potential of success-history based adaptive differential evolution (SHADE)~\citep{tanabe13} and its extension L-SHADE~\citep{tanabe14} to DEBtool. SHADE and L-SHADE are extensions of differential evolution (DE)~\citep{storn97}, one of the most versatile and robust population-based search algorithms of the field of multimodal optimisation~\citep{das2016recent}. Both SHADE and L-SHADE have been successfully applied in areas such as economy~\cite{jana2021}, energy~\citep{biswas2017optimal, gao2021, tran2021}, mathematics~\citep{cantunavila2021}, and pharmacy~\citep{kaur2020}. MultiCalib4DEB offers a wider range of outputs, different calibration solution alternatives, and specific tools to assess parameter uncertainty and validation. MultiCalib4DEB can be directly applied to ongoing studies or existing DEB libraries of species to calibrate the model’s parameters, assess their robustness, and perform visual sensitivity analyses in just one run of the algorithm. Apart from improving the fitting of the models, one of the main advantages of MultiCalib4DEB is to return an optimal and diverse set of equally desirable calibration solutions or point estimates which can be used by the modeller to perform better sensitivity analysis on their results or to launch new calibrations to explore more precisely a specific solution. The application of MMEAs to the calibration of ecological and biological models is not restricted to the DEB theory and therefore, the results of this study and the presented toolbox can be used for similar modeling techniques.

We start by introducing the main concepts in evolutionary computation, niche-preserving techniques and MMEAs. Then, we continue by describing both the main functionalities of MultiCalib4DEB, the parameter calibration process, and the data analysis and the visualisation processes in Section~\ref{sec:tool_description}. Finally, we conclude with the insights coming from the use of MultiCalib4DEB for the calibration of DEB models parameters in Section~\ref{sec:conclusions}. Some examples concerning MultiClib4DEB, such as code samples, figures and descriptions of the calibration options in the toolbox, can be found in the Supporting Information section. In this section, the reader has at his disposal a complete section in which we evaluate the calibration performance of MultiCalib4DEB.

\section{Toolbox description}
\label{sec:tool_description}

\subsection{Main functionalities and DEBtool integration}
\label{subsec:mc4deb_integration}

MultiCalib4DEB is developed to efficiently calibrate DEB models' parameters while returning an optimal set of diverse solutions by using MMEAs~\citep{preuss2015, wong2015}. Due to its population approach, MultiCalib4DEB is able to find both local and global optima of a DEB model while maintaining multiple solutions in a single run. 

The toolbox has been integrated into DEBtool to both improve its performance when calibrating models with a large amount of parameters and to help modellers when analysing the calibration outputs. A set of statistical and visualisation functionalities have been included in MultiCalib4DEB for this purpose. 

MultiCalib4DEB follows the same schema as in DEBtool to easily use, run, save, visualize, and work with the DEB model calibration results. It is thus straightforward and simple to start working with MultiCalib4DEB for someone who is familiar with the DEBtool calibration module. MultiCalib4DEB is fully integrated into DEBtool and the users can use it by setting the option \texttt{`mmea'} into the DEBtool estimation options with \texttt{estim\_options(`method',`mmea');}. The list of modules that integrate MultiCalib4DEB are listed in Table~\ref{table:modules_summarisation}. 

The toolbox is also flexible. Users can edit their calibration settings ranging from the total calibration time to the origin and the ranges of the parameters to calibrate. MultiCalib4DEB provides a wide variety of outputs, different calibration solution alternatives, and specific tools to assess parameter uncertainty and validation which allow users to interact with the calibration process along its execution and also when it finishes. MultiCalib4DEB is coded in MATLAB (v.9.2 (R2017a)).

\subsection{The parameter calibration process}
\label{subsec:params_calibration_options}

\begin{figure*}[ht]
    \centering
    \includegraphics[width=0.8\textwidth]{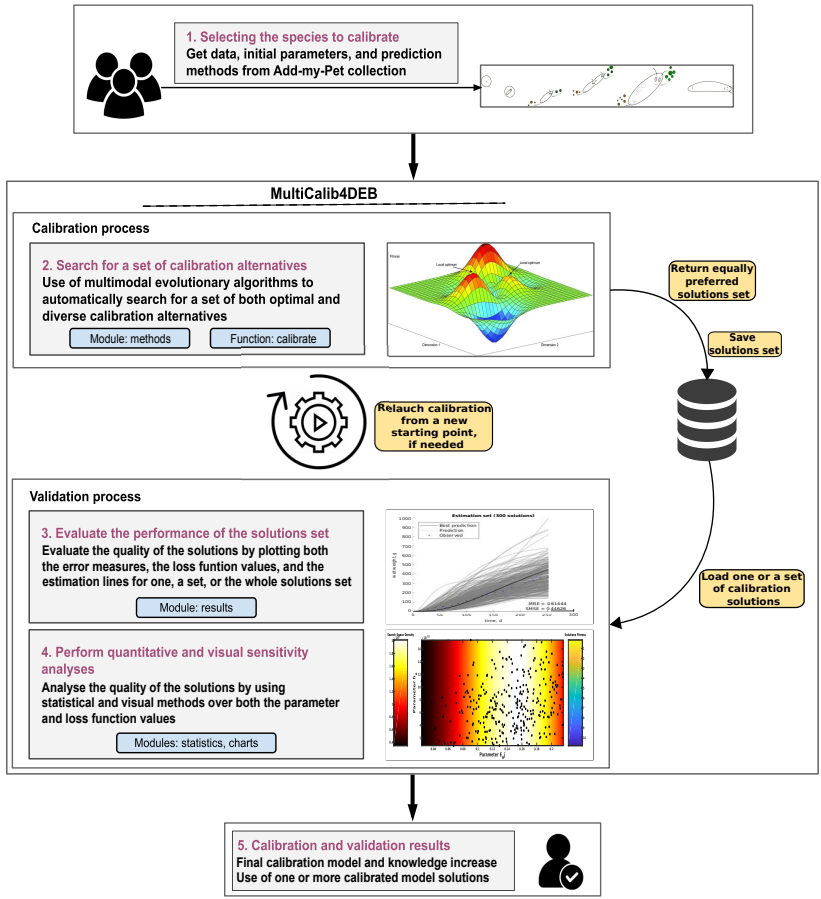}
    \caption{Diagram to illustrate the operation of the MultiCalib4DEB toolbox and its steps during a DEB model calibration process.}
    \label{figure:MC4DEB_diagram}
\end{figure*}

MultiCalib4DEB has been designed to mimic the DEBtool parameter calibration process to facilitate its use by DEBtool users. It is thus possible to launch a calibration process by selecting the species to calibrate, loading the specie data and the DEBtool filters and loss function methods, and configuring some calibration options. After the calibration process, the user can perform both statistical and visual sensitivity analyses to extract insights from the calibration results. MultiCalib4DEB returns different solutions that are equally preferable to each other. These solutions can be saved into a file to be used for further analysis or to initiate a new calibration process to explore a more specific area of a problem search space. In this sense, MultiCalib4DEB does not perform a closed calibration process, but invites to an iterative calibration process in which any of the solutions returned by the toolbox can be used again to launch new calibration processes (e.g., a user can fix some parameter values or modifying their minimum and maximum ranges). Figure~\ref{figure:MC4DEB_diagram} shows a diagram summarizing the calibration process of a DEB model by using MultiCalib4DEB toolbox. 

Although a calibration process can be initiated with a default configuration, the user can customize this configuration with the options in Table S (1) of the Supporting Information section. The user can select from two calibration stopping criterion: 1) the maximum number of evaluations (using the~\texttt{max\_fun\_evals} parameter) and 2) the maximum time (using the~\texttt{max\_calibration\_time} parameter). The calibration stops when a maximum number of evaluations is reached by default. The minimum value recommended for the~\texttt{max\_fun\_evals} parameter is 10,000. A guideline for defining the value of~\texttt{max\_fun\_evals} is to set 1,000 evaluations for each calibration parameter. So, if a species has 15 parameters to calibrate, the maximum number of evaluations should be set at least to 15,000. On the other hand, the minimum calibration time recommended is 1 hour while a suitable criteria for defining the calibration time can be 10 minutes for each calibration parameter. By default, the calibration process stops when one of the above-mentioned criteria is reached, even if convergence is not achieved. A more detailed explanation of the~\texttt{max\_fun\_evals},~\texttt{max\_calibration\_time}, and other calibration options is available in the MultiCalib4DEB's user manual\footnote{\url{https://github.com/JuanfranRobles/MultiCalib4DEB}}. Examples 1 and 2 in the Supporting Information section contains two code examples, showing how to launch different calibration processes for species such as \textit{Clarias Gariepinus} by using different MultiCalib4DEB calibration options.

\subsection{Multimodal evolutionary algorithms}
\label{subsec:mc4deb_algs}

\subsubsection{Main concepts about evolutionary computation and niche-preserving techniques}
\label{sec:evolutionary_niche_preserving}

Solving an optimisation problem involves finding a solution that is the best among others. 
The space of all feasible solutions (the set of solutions among which the desired solution is found) is known as the search space. Each point in the search space represents a possible solution and each possible solution is evaluated by its fitness value for the problem.
 
Evolutionary computation~\citep{back1996evolutionary} provides computational models for search and optimisation that have their origin in evolution theories and Darwinian natural selection. 

In general, evolutionary algorithms use the following scheme for their operation: 

\begin{enumerate}
    \item Adapt a population of candidate solutions to the problem.
    \item Apply a random selective process based on the quality of the generated solutions (measured according to a fitness function).
    \item Alter the selected solutions using crossover and/or mutation operators. 
    \item Use the new solutions generated to replace those of the current population
\end{enumerate}

On the other hand, Niche-preserving techniques are division mechanisms to produce different sub-populations that allow exploring different search space regions (niches) according to the similarity of the individuals~\citep{Goldberg87}.

MMEAs combine the ability of evolutionary algorithms to both explore and exploit the solution space and the capacity of niche-preserving techniques to preserve the necessary diversity between solutions. Thus, MMEAs allow a wide search in different promising regions of the problem search space, avoiding stagnation in sub-optimal solutions and enabling different quality solutions to be obtained in a single run. 

The balance between diversity (different parameter values) and fitness of solutions required by a MMEA represents its main difference with respect to standard evolutionary algorithms, where it is only important to find the best solution in terms of fitness. 

Since the seminal proposal by Goldberg and Richardson~\citep{Goldberg87}, the MMEA family has grown~\citep{kennedy95PSO, dorigo06ACO, tanabe13, tanabe14} and is widely used in a large number of optimisation problems~\citep{das2016recent}.

\subsubsection{Global population-based algorithms}
\label{subsubsec:mmeas}

Two MMEAs are available in the MultiCalib4DEB calibration toolbox: Success-history based adaptive differential evolution (SHADE)~\citep{tanabe13} and its extension L-SHADE~\citep{tanabe14}. Both SHADE and L-SHADE are multimodal extensions of the well-known DE algorithm~\citep{storn97}. 

In addition to their outstanding optimisation performance, SHADE and L-SHADE are adaptive MMEAs. This avoids the need to tune any parameter and makes it possible to use these algorithms without any prior training. The SHADE and L-SHADE algorithms can be briefly described as follows:

\begin{itemize}
    \item \textbf{SHADE}: is a history-based variant of DE in which the \emph{successful} values of the crossover rate probability ($CR$) and mutation rate ($F$) are stored into a historical memory if the solution generated with them improves the previous individual. 
    \item \textbf{L-SHADE}: is an extension of SHADE that incorporates a simple deterministic population resizing method called Linear Population Size Reduction. The resizing method continuously reduces the population size to match a linear function where the population size at first generation is $N_{init}$, and the population at the end of the run is $N_{min}$. 
\end{itemize}

Both SHADE and L-SHADE are extensions of DE, an evolutionary algorithm that generates new solutions by combining existing individuals with a \emph{donor vector}, a mutation rate ($F$), and a crossover rate probability ($CR$) to take the values for the new solution from the \emph{donor vector} or from the original values of the individual. The few parameters SHADE and L-SHADE need for its execution are fixed in MultiCalib4DEB following Tanabe et al. recommendations~\citep{tanabe13, tanabe14}. The value for the crossover rate is set to $CR=0.9$ while the mutation rate is set to $F=0.5$. For both SHADE and L-SHADE we set the minimum size of the historical memory ($P$) to 100. The maximum size for this parameter is defined by the user (for more information about the \texttt{num\_results} parameter, refer to Table S (1) of the Supporting Information section. The $N_{init}$ and the $N_{min}$ for the L-SHADE method are set to the value of $P$ and 5, respectively. Despite L-SHADE reduces the number of solutions through its execution, we maintain the size of the final results set to $P$. Thus, $P$ is the only parameter that is not fixed in MultiCalib4DEB and can be set by the user to control the number of calibration results the toolbox returns after the calibration process. 

\subsubsection{Local refinement of the calibration solutions}
\label{subsubsec:local_search}

To improve the exploitation of the problem search space, the MultiCalib4DEB calibration applies a refinement process over the best, all, or a random set of solutions contained in $P$. The user chooses whether and how to apply the refinement process after calibration with MMEAs. To do so, parameters~\texttt{refine\_best} and~\texttt{refine\_prob} shown in Table S (1) are used. Parameter~\texttt{refine\_best} is the option activating the refinement process on the best solution found after the calibration process while parameter~\texttt{refine\_prob} controls on how many solutions to apply the refinement process. By default, the~\texttt{refine\_best} parameter is activated and the refinement process is applied over the best solution (i.e., those with lower loss function value). The application of a refinement process depends on the desired level of exploration of the problem search space. We consider that applying refinement is a suitable option to search for a global optima in the final solution set. Moreover, this option does not significantly affect the MultiCalib4DEB's execution time as it is restricted to the maximum number of evaluations or the total calibration time defined by the user. 

The application of a refinement process is a novelty with respect to methods such as the one proposed by Filgueira et al.~\citep{Filgueira2020} since, in addition to exploring the solution space in search of different optima, the neighborhood of these optima is also exploited during the search. The refinement process uses a NM simplex method to explore the neighborhood of a local optima to minimize its loss function value. The NM method can use different stopping criteria such as the simplex tolerance (that controls the accuracy of the solution at each simplex iteration) or a convergence criterion (that stops the optimisation process if a minimum convergence threshold is not reached). Following the DEBtool approach, a fixed number of function evaluations is set to search convergence. Then, runs of maximum 500 steps are used for the method, using numerical continuation to restart the optimisation process if convergence is reached. 

\subsection{Data analysis and visualisation}
\label{subsec:data_presentation_and_statistics}

Once the calibration process is finished, the results are stored in a MATLAB object named~\texttt{solutions\_set}. A user can retrieve one or a set of solutions from this object to iteratively re-launch a calibration process with different calibration options if desired. The ~\texttt{solutions\_set} fields are described in Table~\ref{table:results_object_fields}.

Based on the output information of this Table~\ref{table:results_object_fields} the user can also perform both statistical and visual analyses by using the functions in the \texttt{statistics} and the \texttt{charts} modules. The complete list of both statistical and visualisation methods are available in MultiCalib4DEB are listed below: 

\begin{itemize}
    \item \textbf{Statistical functions over loss function values}: cardinality, average, minimum, maximum, standard deviation, and average distance between the loss function set. 
    \item \textbf{Statistical functions over parameter values}: average, standard deviation, spread, minimum, maximum, kurtosis, skewness, bimodal coefficient, and the average distance between the final parameter values and their minimum and maximum ranges. 
    \item \textbf{Visualisation charts}: density heat map (it can be complemented with a scatter plot including the loss function values), scatter plot (simple, weighted by the loss function value, or showing the density of the solutions in the search space), and error plot (calibration values and loss function adjustment over them). 
\end{itemize}

MultiCalib4DEB also allows to visualise and analyse the calibration results with the \texttt{charts} module. This module includes two methods:

\begin{enumerate}
    \item \textbf{\texttt{plot\_chart}}: method to generate different charts (such as scatter plots and heat maps) for pairs of calibration parameters.
    \item \textbf{\texttt{plot\_results}}: method to visualise charts with the real and the predicted values of the calibration results. This method is able to plot either an isolated prediction result for a single calibration solution or a complete report including the prediction values for the whole set of calibration results.
\end{enumerate}

\begin{figure*}[ht]
    \centering
    \begin{subfigure}{\textwidth}
    \hspace{-1cm}
    \includegraphics[width=1.05\textwidth]{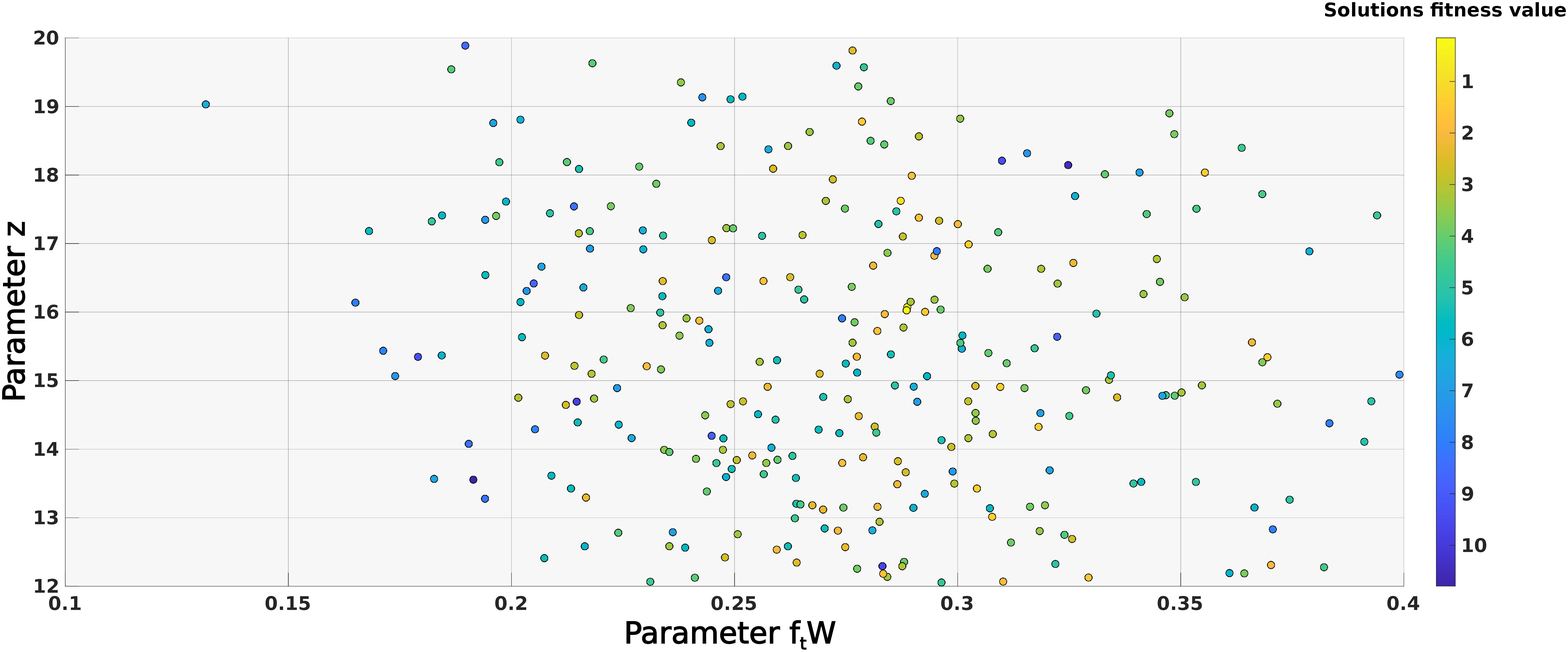}
    \end{subfigure}
    \begin{subfigure}{\textwidth}
    \hspace{0.2cm}
    \includegraphics[width=0.9\textwidth, height=6.5cm]{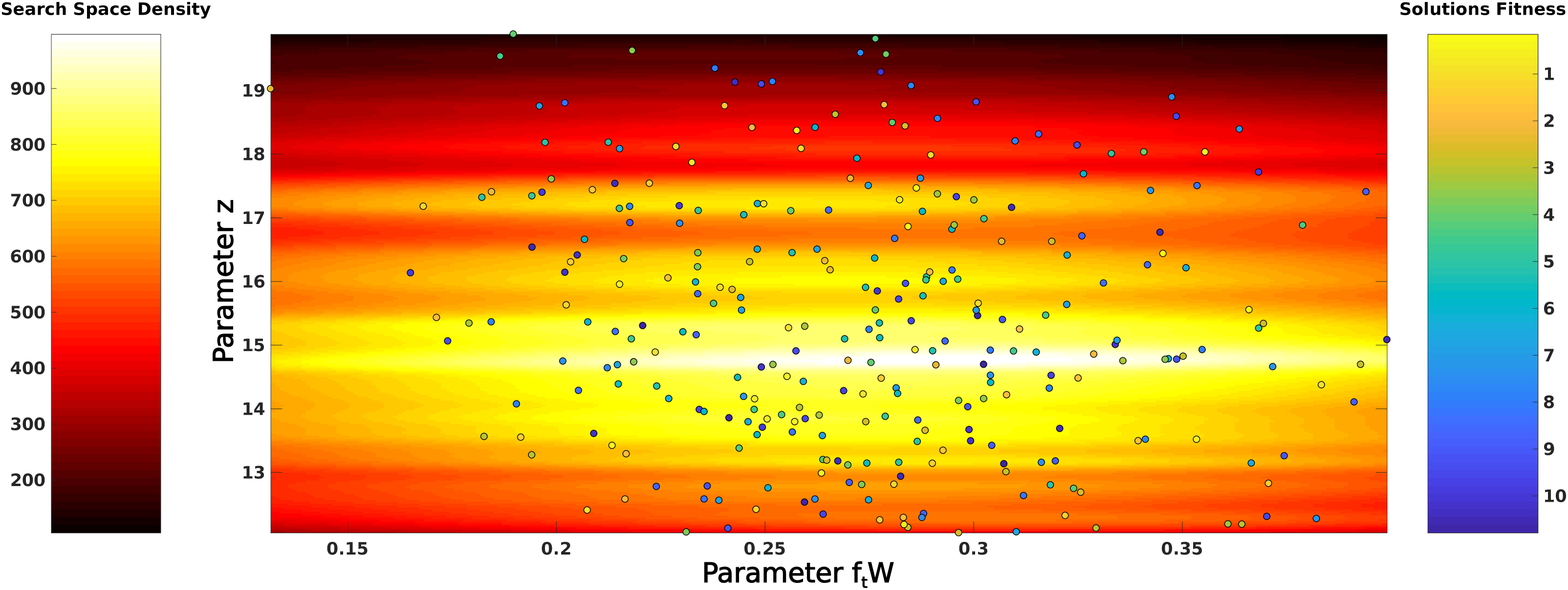}
    \end{subfigure}
\caption {Example of the charts available in the \texttt{charts} module. The figure on the top shows the output from the \texttt{scatter} method while the figure on the bottom shows the output from the \texttt{density\_hm\_scatter} method. The pairs of parameters compared in the charts are the zoom factor ($z$) and $f_{t}W$. $z$ is is defined as the maximum volumetric length of an organism in centimetres while $f_{t}W$ is a scaled functional response for the wet weight at puberty of a species.}
\label{figure:charts_report}
\end{figure*}

\begin{figure*}[ht]
\begin{subfigure}{0.5\textwidth}
\centering\includegraphics[width=\textwidth]{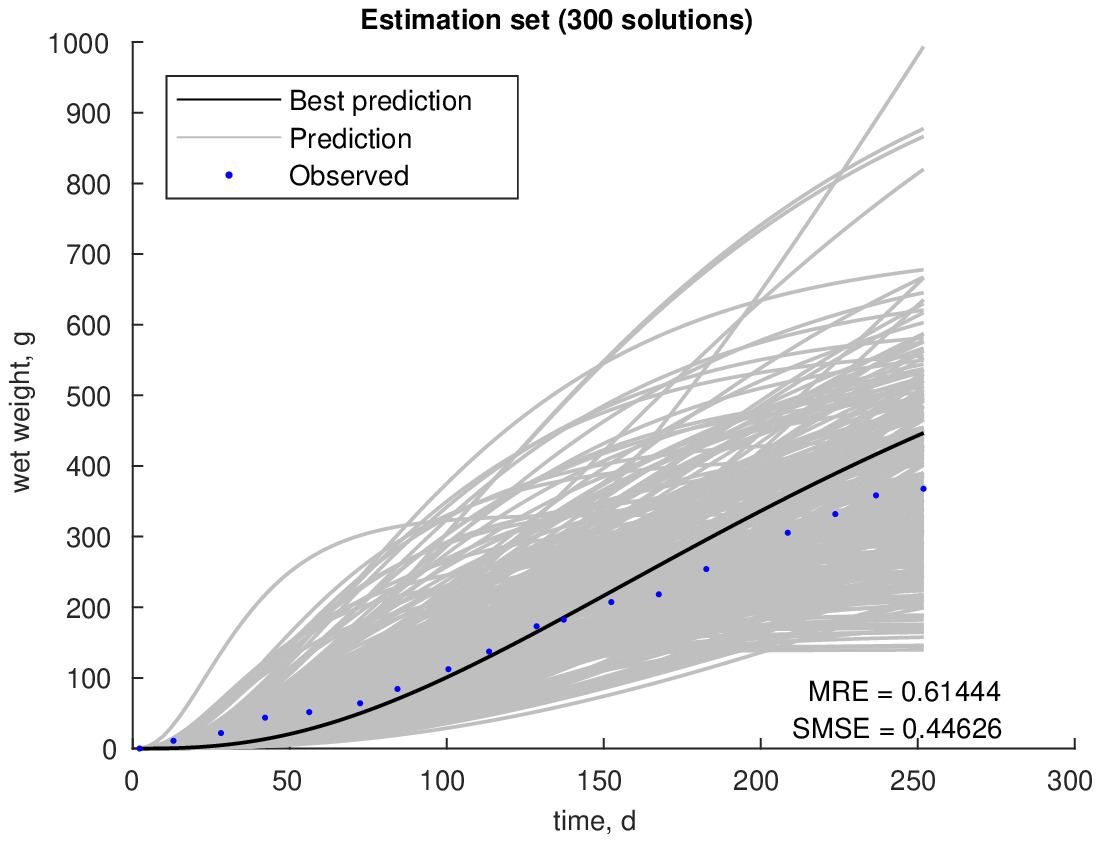}
\end{subfigure}
\hspace{-0.5cm}
\begin{subfigure}{0.5\textwidth}
\centering\includegraphics[width=\textwidth]{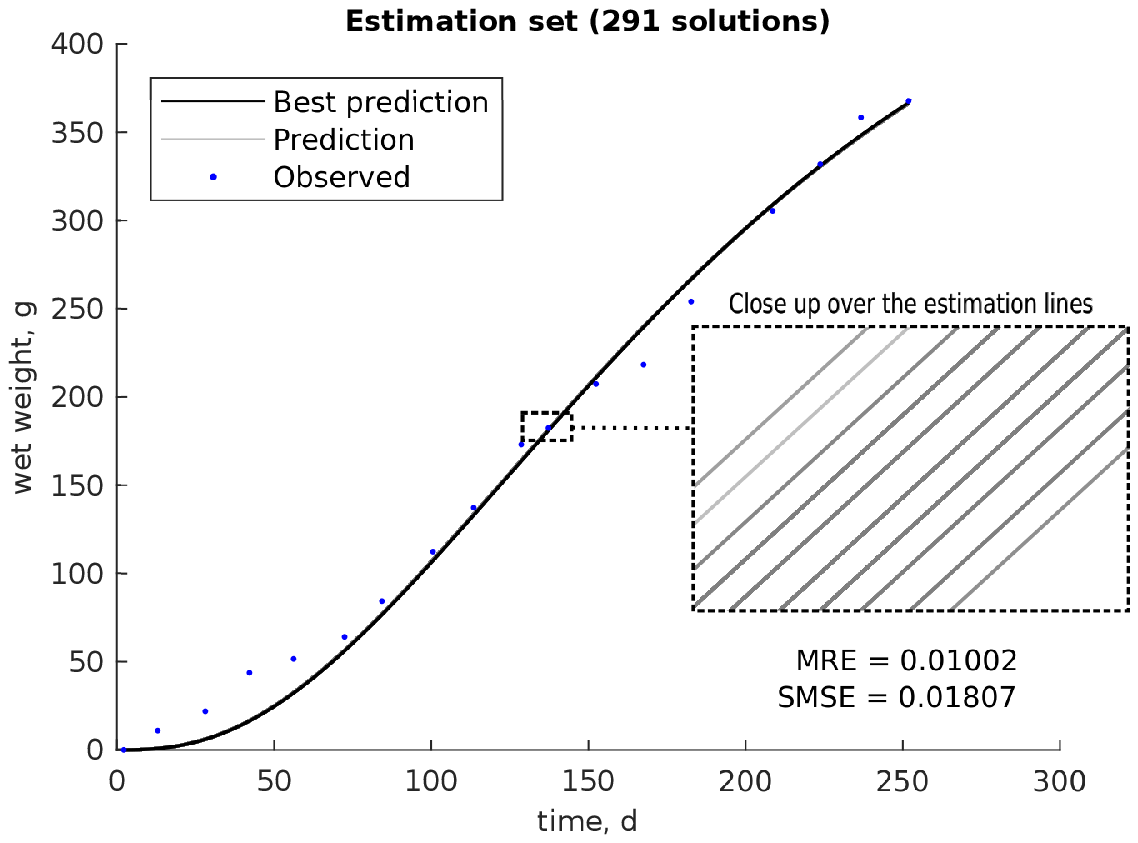}
\end{subfigure}
\caption {Illustrative example for two calibration processes of ten minutes (left) and one hour (right) over the \textit{Clarias gariepinus} species. The X axis contains the time measured in days while the Y axis contains the values of the wet weight (in grams). The figure shows a summary of the set of solutions for the two calibration processes where the blue circles represent the real observed values, the grey lines the prediction values for every solution in the final solutions set, and the black line, the result with the lowest (best) loss function value. In the one hour calibration, the estimation of all solutions is overlapped, as can be seen with the zoom in panel.}
\label{figure:calibration_example}
\end{figure*}

We list the options that are available for both the \texttt{plot\_chart} and \texttt{plot\_results} methods in Table~\ref{figure:plot_options}. With the \texttt{plot\_chart} method, an user can explore the relationships between a pair of calibration parameters in the search space. The different charts this method offers can be useful to better understand how the parameters are related and the search space they explore. Figure~\ref{figure:charts_report} describes the plots that can be generated with the \texttt{plot\_chart} method while Figure~\ref{figure:calibration_example} shows plots that can be generated with the \texttt{plot\_results} method. The options in the \texttt{plot\_results} method can be used to represent either the loss function values obtained or the errors. Figure~\ref{figure:charts_report} shows an illustrative example of the output obtained with this method.

Examples 3 and 4 in the Supporting Information section show two examples with code function calls to some of the methods in Table~\ref{figure:plot_options}. The results obtained from Examples 3 and 4 are illustrated in Figure S (1) of the Supporting Information section.

\section{Evaluation of the calibration performance}
\label{sec:calibration_outputs}

MultiCalib4DEB returns different results through the calibration process. Thus, both the quality of the solutions that the calibration engine finds and their diversity increase over time. Figure~\ref{figure:calibration_example} shows two examples of calibration results from two calibration processes running for 10 minutes (left panel) and 1 hour (right panel) respectively when calibrating the \textit{Clarias gariepinus} species base model of the abj DEB model. Figure~\ref{figure:calibration_example} compares the wet weight (measured in grams), which measures the mass of a species, against time (measured in days), respectively. As can be observed, the quality of the solutions and the error measures is considerably better in the one hour calibration process compared to the ten minutes one because, in this case, the algorithm does not have time to explore the problem search space and to converge. Within the final set of solutions of the 1 hour calibration process, there is one that has the lowest (best) loss function value. The remaining solutions have higher loss function values than the best solution, but have accurate predictions and are equally desirable. While numerical methods such as NM are limited to providing a single solution, MMEAs generate a set of solutions that could allow users to explore the parameters of the DEB and its behaviour.

A further numerical exercise is carried out to compare the performance of NM and SHADE with refinement calibration algorithms by calibrating 13 different species from DEB's laboratory species repository\footnote{\url{http://www.bio.vu.nl/thb/deb/deblab/add_my_pet/species_list}}. The match between data and predictions is quantified by the goodness of fit using the loss function value, a weighted mean absolute percentage error (MAPE) which is referred to as mean relative error (MRE) in the DEB literature, and the symmetric mean squared error (SMSE)~\citep{marques2019}. MRE evaluates the differences between the data and the predictions in an additive way, while SMSE does it in a multiplicative way. The MRE and the SMSE are computed as follows:

\begin{equation}
\begin{split}
	MRE = \frac{1}{n'} \sum_{i=1}^{n} \sum_{j=1}^{n_{i}} \frac{w_{ij}}{w_{i}} \frac{|p_{ij}-d_{ij}|}{|d_{i}|},  \\ \mbox{where }  w_{i}=\sum_{j=1}^{n_{i}} w_{ij} > 0 \mbox{ and } d_{i}=\frac{1}{n_{i}} \sum_{j=1}^{n_{i}}d_{ij}
\label{equation:MRE}
\end{split}
\end{equation} 
\begin{equation}
\begin{split}
	SMSE =  \frac{1}{n'} \sum_{i=1}^{n} \sum_{j=1}^{n_{i}} \frac{w_{ij}}{w_{i}} \frac{(p_{ij}-d_{ij})^2}{(p_{i}^2+d_{i}^2)}, \\ \mbox{where } w_{i}=\sum_{j=1}^{n_{i}} w_{ij} > 0 \mbox{ and } p_{i}=\frac{1}{n_{i}} \sum_{j=1}^{n_{i}}p_{ij}
\label{equation:SMSE}
\end{split}
\end{equation} 
where $i$ refers to the species data set and $j$ to a given point in $i$. $d_{ij}$ stands for the data, $p_{ij}$ for the model prediction, $w_{ij}$ for the associated weight coefficient. $n'$ is the number of data sets with $w_{i} > 0$. The weight coefficients quantify the confidence of the user in the data sets as well as for specific data point. They are automatically set to $w_{ij}=\frac{1}{n_{i}}$ until the user overwrite them. MRE takes values from 0 to infinity, while SMSE takes values from 0 to 1. In both cases, 0 means that the predictions exactly match the data. 

In order to perform a fair comparison of both calibration algorithms, the experimental design considers 20,000 evaluations of the loss function as a stopping criterion (15,000 for SHADE and runs of maximum 500 steps for the refinement process by using NM). When SHADE finishes its search across the search space, the refinement process is applied over the best solution found. The refinement runs until the maximum number of evaluations is reached or the loss function no longer improves. The single NM method follows the same stopping criterion as the one used in SHADE (i.e., 20,000 evaluations). The NM method is iteratively launched with runs of maximum 500 steps (i.e., 500 evaluations of the loss function) following the estimation procedure recommendations\footnote{\url{http://www.debtheory.org/wiki/index.php?title=AmP_estimation_procedure}}. Thus, when the method reaches 500 steps, the search continues from the previously (best) obtained result. When 20,000 evaluations are accomplished or there is not improvement, the method stops. The calibration options are set by default for the DEBtool and MultiCalib4DEB calibration engines. Then, the number of solutions to maintain during the calibration with SHADE ($P$) is set to 200. Thus, a set of 200 optimal solutions is returned after the calibration process for each calibrated species. Both SHADE and NM start the calibration with the initial species parameters defined in Add-my-Pet project database. 

As can be observed in Table~\ref{table:calibration_results_comparison}, the initial loss function values are usually close to the minimum loss function values returned by the algorithms as sometimes the calibration starts in the neighborhood of the best solution. The improvement between NM and SHADE is computed as the percentage increases of the SHADE minimum (best) loss function values against the NM ones. A value of zero means that SHADE does not improve the NM results while a value greater than zero means that SHADE outperforms Nelde Mead (e.g. a value of 0.05 means a 5\% improvement of SHADE over NM). The complete list of the species we select is shown in Table~\ref{table:calibration_results_comparison} including their typified model\tnote{A typified model is the class of the species that DEB theory allows for the construction of DEB models. Each typified model corresponds to a similar species or taxonomic group (consult \url{http://www.debtheory.org/wiki/index.php?title=Typified_models} for further information about typified models.}, the number of parameters to calibrate, and the initial value of the loss function.

Table~\ref{table:calibration_results_comparison} shows that SHADE is able to successfully calibrate species with different complexities. SHADE outperforms NM according to the final loss function value in 9 out of 13 species while NM achieves the same best final loss function value as SHADE in four species. The SHADE algorithm is able to reduce the loss function value of the calibrated species between one (e.g. for the \textit{Dipodomys herrmanni}, \textit{D. merriami}, and \textit{Lepus timidus} species) and four percent (e.g. for the \textit{Dipodomys deserti} and \textit{Magallana gigas} species) with respect to the NM results. Moreover, the SHADE algorithm returns 200 optimal and diverse solutions for each species instead of the single solution returned by NM. Additionally, the loss function average values for the 200 solution set of each species is close to the minimum loss function values, showing that any of the solutions in the final solutions is optimal and can be considered as a feasible calibration solution. SHADE also achieves the best SMSE and MRE values in seven species. Furthermore, SHADE outperforms NM by tackling the most complex species in terms of number of parameters (\textit{Magallana gigas}, 18 parameters to calibrate) according to the final loss function value, SMSE and MRE.

\section{Conclusions and further research}
\label{sec:conclusions}

MultiCalib4DEB aims to help researchers to better calibrate the parameters of DEB models. The toolbox uses powerful MMEAs to calibrate DEB models. 
The strength of the use of those algorithms is two-fold. On the one hand, they have demonstrated their outstanding performance in a wide range of application fields. On the other hand, they are specially suitable tackling problems that include a set of optimal solutions. MMEAs are designed to return that set of optimal solutions in a single run. We perform the calibration of 13 species with different complexities to compare SHADE against NM. The results of the experimentation shows a performance between a 1\% and a 4\% greater of SHADE with respect to the NM algorithm while returning 200 optimal and diverse solutions for each species instead of a a single solution. From this set of solutions, the user can choose which ones to visualise, analyse, or use for another calibration process. This last statement reinforces the benefits of using MMEAs for the calibration of DEB models. 

The MultiCalib4DEB toolbox is also able to perform statistical analyses and to generate visualisation charts. These tools are useful to explore and analyse the relationships among the calibration parameters. Both statistical and visualisation modules can be used to validate the parameter calibration results. MultiCalib4DEB is particularly suitable for DEBtool users who aim to perform sensitivity analyses from different calibration results. 

MultiCalib4DEB is specially helpful to efficiently calibrate different species independently of the amount of calibration parameters while returning several parameter configurations in a single run. The toolbox is also flexible to be adapted to the user needs. It has been developed to enhance the calibration of DEB models within DEBtool but it is not limited to the calibration of DEB models.


\begin{table*}
\caption{Description of the MultiCalib4DEB modules}
\label{table:modules_summarisation}
\begin{threeparttable}
\begin{tabular}{lp{.8\textwidth}}
\headrow
\textbf{Module name} & \textbf{Description}\\
\texttt{charts} & Multiple plot options for the validation of the MultiCalib4DEB calibration results such as heatmaps and scatter plots \\
\texttt{configuration} & File that controls the calibration options \\
\texttt{examples} & Examples for the calibration of different species by using the MultiCalib4DEB calibration algorithms and options. It also contains examples for the visualisation and statistical report modules \\
\texttt{functions} & Auxiliary functions for the calibration algorithms, plots, and statistics \\
\texttt{methods} & Code of the SHADE, L-SHADE, and the local search algorithm which are used for calibration into MultiCalib4DEB \\
\texttt{results} & Functions that generate the plots from the calibration results. \\
\texttt{statistics} & Functions that generate the statistical reports from the calibration results \\
\texttt{utils} & Utilities to save and generate solution reports \\
\hiderowcolors
\hline  
\end{tabular}
\end{threeparttable}
\end{table*}

\begin{table*}
\caption{MultiCalib4DEB output: Description of the information provided after the calibration process.}
\label{table:results_object_fields}
\begin{threeparttable}
\begin{tabular}{l p{.8\textwidth}}
\headrow
\textbf{Field name} & \textbf{Description}\\
\texttt{set\_size} & Number of solutions returned after the calibration process\\
\texttt{solutions\_set} & Set of solutions returned after the calibration process. It contains \texttt{set\_size} solutions, each of them with a different set of calibration parameters \\
\texttt{fun\_values} & Loss function values for each solution in \texttt{solutions\_set}\\
\texttt{par\_names} & List of parameters that are selected for calibration. It is an information field that is used in the execution of the \texttt{statistics}, \texttt{results} and \texttt{charts} modules. \\
\texttt{results} & MatLab struct field that contains both general information about the species whose parameters are calibrated and specific information about each solution in $solutions\_set$. All the calibration solutions are listed in this field as subfields with the name ``solution\_ \texttt{+} solution\_number'' and they include the ``par'' and ``metaPar'' files with the species parameters in DEBtool format. Then, the general information is stored below the solutions list and contains the ``data'', ``auxData'', ``txtPar'', ``metaData'', ``txtData'', and ``weights'' also in DEBtool format. The information in this field facilitates working with one or a set of results after the calibration process. In addition, the solutions in the texttt{results} field can be used to generate reports later by using the texttt{results} module of MultiCalib4DEB \\
\hiderowcolors
\hline  
\end{tabular}
\end{threeparttable}
\end{table*}

\begin{table*}
\centering
\caption{Options for plotting results and charts from MultiCalib4DEB output.}
\label{figure:plot_options}
\begin{threeparttable}
\begin{tabular}{l p{.2\textwidth} p{.6\textwidth}}
\headrow
\textbf{Method} & \textbf{Option} & \textbf{Description}\\
                  & \texttt{density\_hm} & Plots a heat map with the distribution of the loss function values from the MultiCalib4DEB results in a two-dimensional search space. The final heat map is plotted for a pair of calibration parameters \\
                  & \texttt{density\_hm\_scatter} & Plots a heat map together with a scatter plot in which the values for the pair of parameters selected are represented  \\
\multirow{1}{*}{plot\_chart} & \texttt{scatter} & Plots a scatter plot with the values for a pair of parameters from the calibration results returned by MultiCalib4DEB \\
                  & \texttt{weighted\_scatter} & Plots a scatter plot in which the parameter values are weighted by using its loss function value \\
                  & \texttt{density\_scatter} & Plots a scatter plot in which each point is colored with the spatial density of nearby points. The function uses the kernel smoothing function to compute the density value for each point \\
                  & \texttt{prediction} & Plots a prediction plot, the default plot the DEBtool toolbox returns after a calibration process  \\
                  \hline
\multirow{6}{*}{plot\_results} & \texttt{Basic} & Plots prediction results \\
                  & \texttt{Best} & Plots the calibration from the parameters of the best solution \\
                  & \texttt{Set} & Plots a grouped calibration for the whole results set together with the average MRE and SMSE \\
                  & \texttt{Complete} & Plots \texttt{Basic}, \texttt{Best}, and \texttt{Set} options \\
\hiderowcolors
\hline  
\end{tabular}%
\end{threeparttable}
\end{table*}

\begin{table*}
\begin{minipage}{\textwidth}
\centering
\scriptsize
\caption{Comparison between NM and SHADE calibration results for different species. The table contains the number of parameters to calibrate and the initial loss function for each species. The minimum (best) loss function, SMSE and MRE values are also included both for the NM and the SHADE algorithms. The average loss function from the 200 solutions SHADE returns are shown to the rigth of the best loss function values. The best values for the loss function, the SMSE, and the MRE are in bold.}
\label{table:calibration_results_comparison}
\begin{threeparttable}
\begin{tabular}{P{0.18\textwidth}P{0.06\textwidth}P{0.06\textwidth}P{0.07\textwidth}ccP{0.11\textwidth}P{0.08\textwidth}ccc}
\hiderowcolors
\rowcolor{Gray}
& & & \multicolumn{3}{c}{\textbf{NM (1 solution)}} & \multicolumn{4}{c}{\textbf{SHADE (set of 200 solutions)}} \\
\rowcolor{Gray}
\multirow{+3}{*}{\textbf{Species (Typified model)}} & \textbf{Parameters to calibrate} & \multirow{+3}{*}{\textbf{\makecell[c]{Initial loss \\ function}}} & \textbf{Best final loss function} & \multirow{+3}{*}{\textbf{SMSE}} & \multirow{+3}{*}{\textbf{MRE}} & \textbf{Best final loss function (Improvement)} & \textbf{Average loss function of the set} & \multirow{+3}{*}{\textbf{SMSE}} & \multirow{+3}{*}{\textbf{MRE}} \\ \hline
\hiderowcolors
\textit{Cyclops vicinus} (abp) & 9 & 6.3819 & 0.4848 & 0.1714 & 0.198 & 0.4848 (0.0) & 0.485 & \textbf{0.1713} & \textbf{0.1974} \\
\textit{Lobatus gigas} (abj) & 9 &  0.0623 & 0.0623 & 0.0381 & 0.0365 & 0.0623 (0.0) & 0.0623 & \textbf{0.0378} & \textbf{0.0364} \\
\textit{Dipodomys deserti} (stx) & 10 & 0.2326 & 0.2326 & 0.021 & 0.0189 & \textbf{0.2234 (0.04)} & 0.2266 & \textbf{0.0185} & \textbf{0.0169} \\
\textit{Dipodomys herrmanni} (stx) & 10 & 0.3289 & 0.3289 & \textbf{0.042} & 0.024 & \textbf{0.3237 (0.016)} & 0.3241 & 0.0439 & \textbf{0.0228} \\
\textit{Dipodomys merriami} (stx) & 10 & 0.3167 & 0.3135 & 0.018 & 0.017 & \textbf{0.3102 (0.011)} & 0.3113 & \textbf{0.0165} & \textbf{0.0154} \\
\textit{Lepus timidus} (stx) & 10 & 0.3068 & 0.3066 & 0.013 & 0.013 & \textbf{0.3028 (0.0124)} & 0.303 & \textbf{0.0075} & \textbf{0.0073} \\
\textit{Clarias gariepinus} (abj) & 11 & 0.1531 & 0.15293 & 0.019 & 0.013 & \textbf{0.1516 (0.009)} & 0.1525 & \textbf{0.0181} & \textbf{0.011} \\
\textit{Heterobranchus longifilis} (abj) & 11 & 0.12205 & 0.1183 & 0.0161 & 0.0131 & 0.1183 (0.0) & 0.1404 & 0.0161 & 0.0131 \\
\textit{Porcellio scaber} (std) & 12 & 0.1887 & 0.0988 & 0.0472 & 0.0522 & 0.0988 (0.0) & 0.1294 & 0.0472 & 0.0522 \\
\textit{Pleurobrachia bachei} (abj) & 12 & 1.3614 & 1.3503 & 0.24 & \textbf{0.182} & \textbf{1.3174 (0.0244)} & 1.3182 & \textbf{0.2364} & 0.1838 \\
\textit{Homo sapiens} (stx) & 12 & 0.28 & 0.28 & 0.057 & 0.051 & \textbf{0.2794 (0.0022)} & 0.2794 & \textbf{0.0547} & \textbf{0.0495} \\
\textit{Asterias rubens} (abj) & 15 & 0.98 & 0.8443 & 0.1431 & 0.1432 & \textbf{0.8442 (0.0002)} & 0.8442 & \textbf{0.143} & \textbf{0.1431} \\
\textit{Magallana gigas} (asj) & 18 & 16.7519 & 15.527 & 0.463 & 0.405 & \textbf{14.8837 (0.0414)} & 15.0611 & \textbf{0.4553} & \textbf{0.3965} \\
\hline  
\end{tabular}
\end{threeparttable}
\end{minipage}
\end{table*}

\newpage 

\ 

\newpage

\end{document}


\setlength{\parindent}{0ex}
\textbf{SUPPORTING INFORMATION} 
\\\\
APPENDIX 1: THE MULTICALIB4DEB CALIBRATION OPTIONS
\\

\begin{table}[h]
\begin{minipage}{\textwidth}
\centering
\scriptsize
\caption{Description of the options for parameter calibration}
\label{table:calibration_options}
\begin{threeparttable}
\begin{tabular}{p{.12\textwidth} p{.5\textwidth} p{.25\textwidth}}
\headrow
\textbf{Name} & \textbf{Description} & \textbf{Values} \\
\texttt{method} & Sets the MMEA to use for calibration & \makecell[cl]{mm1: SHADE \\ mm2: L-SHADE}  \\
\texttt{num\_results}\tnote{1} & Defines the number of individuals in the SHADE and L-SHADE populations that will be returned after the calibration process as calibration results & $[50, 500]$ \\
\texttt{gen\_factor}\tnote{2} & It is a percentage value used to construct the ranges from which to generate the first set of solutions & $(0.0, 1.0)$ \\
\texttt{bounds\_from\_ind} & It is a parameter that controls where the parameters for the initial population of individuals are taken from & \makecell[l]{0: from pseudo data values \\ 1: from initial data values} \\
\texttt{add\_initial} & Controls whether the species parameters (those existing species parameters from the Add-my-Pet project) are entered in the first population of the MMEA & \makecell[cl]{0: no \\ 1: yes} \\ 
\texttt{refine\_best} & Controls if the best solution is refined by using a local search procedure after the calibration process & \makecell[cl]{0: no \\ 1: yes} \\
\texttt{refine\_prob} & It is a probability value that controls on which solutions to apply the refinement process after calibration with MMEAs\tnote{4} & \makecell[cl]{[0.0, 1.0]} \\
\texttt{max\_fun\_evals} & Defines the maximum number of function evaluations for the calibration process & $[10,000, 100,000]$ \\
\texttt{max\_calibration\_time} & Defines the maximum calibration time in minutes for the calibration process & [60, 10,080 (one week)] \\
\texttt{num\_runs} & Defines the number of independent runs to perform through the calibration process. Each run has an independent seed & [1, 15] \\
\texttt{verbose} & Controls whether to print information on the best parameters found through the calibration process & \makecell[cl]{0: no \\ 1: yes} \\
\texttt{verbose\_options}\tnote{4} & Defines the number of calibration solutions to be printed. This parameter is only used when the \texttt{verbose} option is enabled. & $[10, num\_results]$ \\
\texttt{ranges} & Allows to define a set of ranges for all or a subset of the parameters to be calibrated. This parameter receives a parameter-range or parameter-percentage pair to define the minimum and maximum values of the parameter(s) & \makecell[cl]{$[min_{param}, max_{param}]$\tnote{5} \phantom{ }or \% value\tnote{6} } \\ 
\texttt{results\_output} & It is an option to print the results after the parameter calibration process & \makecell[l]{Basic: It does not show the results on \\ the screen but a summary of the best \\ result in text mode\\ Best: Plots the prediction of the best \\ result by using the DEBtool style \\ Set: Plots the prediction of the whole \\ result set by using the DEBtool style \\ } \\  
\texttt{results\_filename} & Defines the name of the results file & \makecell[l]{If empty or undefined, the name of the \\ result file is generated by default. \\Otherwise, the result file gets the name \\that is set into this field. } \\
\hline  
\end{tabular}
\begin{tablenotes}
\item $^1$ The greater the value of this parameter is, the slower the calibration process is. 
\item $^2$ A value of 0.9 means that, for an initial parameter value of 1.0, the range for generation is $[(1 - 0.9) \cdot 1.0, 1.0 \cdot (1 + 0.9)]$. Thus, the generated values will be selected randomly from the range [0.1, 1.9].
\item $^4$ A value of 0 means that no refinement is applied while a value of 1 means that a refinement is applied over all the solutions. If the value of this parameter is 0.05, the refinement process is applied over the 5\% of the solutions. 
\item $^4$ It is preferable to set a low value for this parameter to better display the best parameters that the calibration algorithms find through the calibration process.
\item $^5$ When a range [min., max.] is defined for a parameter, the values generated through the calibration process cannot underpass or overpass the values set as minimum and maximum in the range. When an calibrated value is under the minimum its value turns to the minimum value set in the range. When the value is over the maximum then it changes to the maximum value defined in the parameter's range.  
\item $^6$ When a percentage is defined for a parameter, the minimum and maximum ranges for these parameter are calculated as $[min. = parameter\_value \cdot (1.0 - \% value), max. = parameter\_value \cdot (1.0 + \% value)]$ where $\% value$ is the factor that is applied to the original parameter value to generate the parameter ranges used to initialize the first population of individuals. 
\item 
\end{tablenotes}
\end{threeparttable}
\end{minipage}
\end{table}

\newpage

APPENDIX 2: CALIBRATION EXAMPLES AND OUTPUTS
\begin{itemize}
    \item \textbf{Example 1}: Calibrate parameters for \textit{Clarias gariepinus} species with SHADE by taking the initial parameters values from the generalised organisms, adding the initial parameters into algorithm's initialization, running one single algorithm's run, without ranges for the free parameters being calibrated, and 10,000 evaluations as the stopping criterion. Verbose is not activated. 
    
    \begin{footnotesize}
    \begin{verbatim}
close all; 
global pets

pets = {'Clarias_gariepinus'}; % The species to calibrate
check_my_pet(pets); % Check species consistence

% Default DEBtool_M options (filters, loss function, ...)
estim_options('default'); 

calibration_options('default'); % Setting default calibration options 
calibration_options('method', 'mm1'); % Calibrate with SHADE
calibration_options('max_fun_evals', 10000); % Stop in 10,000 evaluations
% Take the parameters from pseudo data
calibration_options('bounds_from_ind', 1); 
% Add the initial values into SHADE initialization
calibration_options('add_initial', 1); 
calibration_options('verbose', 0); % Deactivate verbose
[best, info, out, best_fvalues] = calibrate; % Calibrate!
    \end{verbatim}
    \end{footnotesize}
    
    \item \textbf{Example 2}: Calibrate parameters with L-SHADE using the initial parameters values from the species pars\_init file\footnote{\url{http://www.bio.vu.nl/thb/deb/deblab/add_my_pet/entries_web/Clarias_gariepinus/Clarias_gariepinus_res.html}}. Then, generate the population from the species values by applying a generation factor of 20\% over the parameters base values, do not add the initial parameter values to algorithm's initialization, and run five different algorithm's runs. Finalize the calibration process by refining the best solution. The criterion to stop the calibration is to achieve 30 minutes. The verbose option is activated and it prints the ten best function values on screen. 
    
\begin{footnotesize}
\begin{verbatim}
close all; 
global pets

pets = {'Clarias_gariepinus'}; % The species to calibrate
check_my_pet(pets); % Check species consistence

% Default DEBtool_M options (filters, loss function, ...)
estim_options('default');

calibration_options('default'); % Setting default calibration options
calibration_options('method', 'mm2'); % Calibrate with L-SHADE
% Stop in after 30 minutes of calibration
calibration_options('max_calibration_time', 30); 
% Take the parameters from species data
calibration_options('bounds_from_ind', 0); 
% Set value for parameter generation ranges
calibration_options('gen_factor', 0.2); 
% Do not add the initial values into SHADE initialization
calibration_options('add_initial', 0); 
% Run the calibration five times (with different random seeds)
calibration_options('num_runs', 5); 
calibration_options('refine_best', 1); % Refine the best solution
calibration_options('verbose', 1); % Deactivate verbose
% Print the best ten function values on screen
calibration_options('verbose_options', 10); 
[best, info, out, best_fvalues] = calibrate;  % Calibrate!
\end{verbatim}
\end{footnotesize}

\newpage

    \item \textbf{Example 3}: Launching \texttt{plot\_chart} in MultiCalib4DEB:
    
\begin{footnotesize}
\begin{verbatim}
global pets 

% The species to calibrate
pets = {'Clarias_gariepinus'};
% Check pet consistence
check_my_pet(pets);

[data, auxData, metaData, txtData, weights] = mydata_pets; % Get species data

% Load the solution set (example for Clarias Gariepinus). 
load('solutionSet_Clarias_gariepinus_20-Apr-2021_20:42:00.mat');

% Plot the chart!
plot_chart(solutions_set, 'density_hm', {'kap'; 'E_G'}, true, 20);
\end{verbatim}
\end{footnotesize}

    \item \textbf{Example 4}: Launching \texttt{plot\_results} in MultiCalib4DEB:
    
\begin{footnotesize}
\begin{verbatim}
global pets 

pets = {'Clarias_gariepinus'}; % The species to calibrate
check_my_pet(pets); % Check species consistence

% Get species data
[data, auxData, metaData, txtData, weights] = mydata_pets;
https://www.overleaf.com/project/60620c957d40a9dd3909e2fa
% Load the solution set (example for Clarias Gariepinus). 
load('solutionSet_Clarias_gariepinus_20-Apr-2021_20:42:00.mat')

% Plot the solutions!
plot_results(solutions_set, solutions_set.results.txtPar, ...,
             solutions_set.results.data, ...,
             solutions_set.results.auxData, metaData, ..., 
             solutions_set.results.txtData, weights, 'Set');
\end{verbatim}
\end{footnotesize}

\end{itemize}

\begin{figure}[ht]
\begin{subfigure}{\textwidth}
\centering\includegraphics[width=0.8\textwidth]{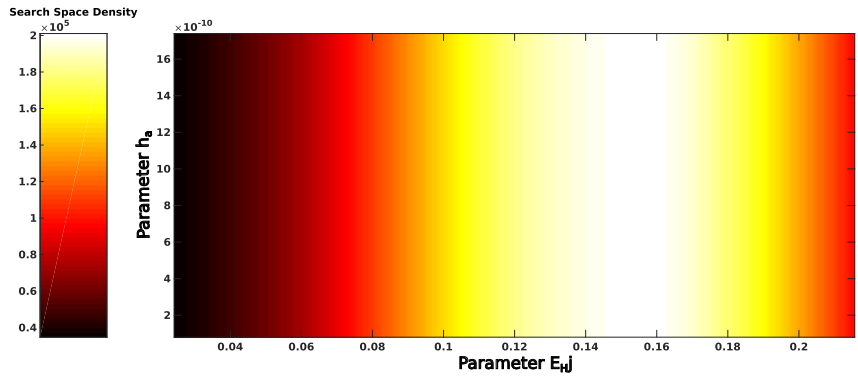}
\end{subfigure}
\begin{subfigure}{\textwidth}
\centering\includegraphics[width=0.7\textwidth]{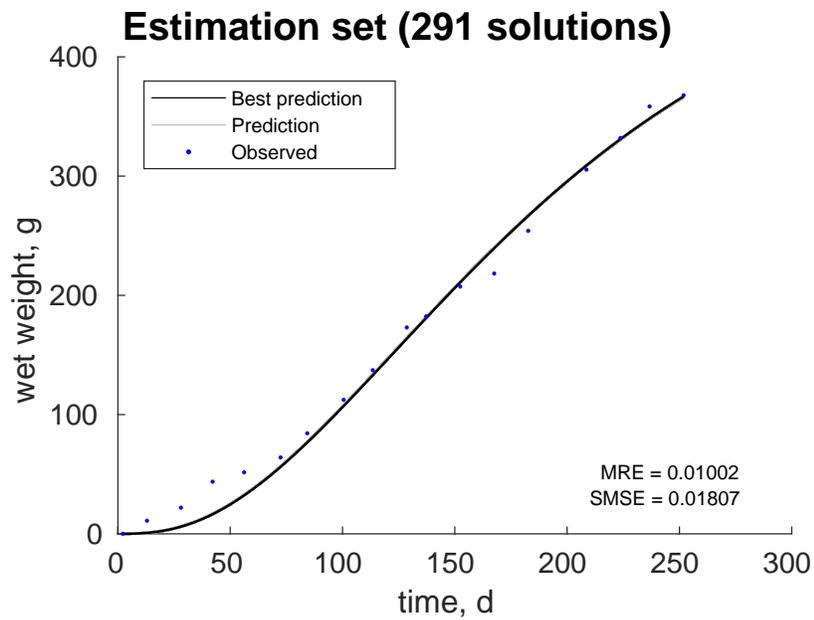}
\end{subfigure}
\caption {Example of the output from the \texttt{charts} and \texttt{plots} modules that MultiCalib4DEB brings. The plot on the top shows the output from Example 3 while the plot on the bottom shows the output from Example 4.} 
\label{figure:plot_and_chart_modules_example}
\end{figure}